\ifdefined\pdfminorversion
\fi
\documentclass[11pt]{article}

\usepackage[preprint]{acl}

\usepackage{times}
\usepackage{latexsym}

\usepackage[T1]{fontenc}
\usepackage[utf8]{inputenc}

\usepackage{microtype}

\usepackage{inconsolata}

\usepackage{graphicx}

\usepackage{booktabs}
\usepackage{multirow}
\usepackage{colortbl}
\usepackage{url}
\usepackage{amsmath}    % 提供\text{}命令，支持数学环境中的文本
\usepackage{amssymb}    % 提供\mathbb{}命令，支持黑板粗体符号
\usepackage{makecell} % 加载makecell宏包（必须）
\usepackage{amsfonts}
\usepackage{tcolorbox}
\usepackage[nospread,noshrink]{flushend}
\definecolor{softbluegray}{RGB}{226,232,238} % 柔和的蓝灰色
\newtcolorbox{takeaway}[1]{
    colback=softbluegray,    % 背景颜色（浅灰色）
    colframe=softbluegray,  % 边框颜色（深灰色）
    fonttitle=\bfseries,      % 标题字体加粗
    coltitle=black,           % 标题颜色
    attach title to upper,    % 将标题直接放在内容上方（不使用单独的标题背景条）
    after title={\ },         % 标题后的间隔
    sharp corners,            % 直角边框
    boxrule=0.pt,             % 边框粗细
    left=5pt, right=5pt, top=5pt, bottom=5pt, % 内部边距
    title={#1},               % 传入的参数作为标题
}

\title{HARPO: Hallucination-Aware Reinforcement Learning for Faithful and Creative Language Generation}

\author{Tiezheng Yu, Yuxin Jiang, Jinpeng Li, Shuning Sun \\
  \textbf{Fei Mi, Haoli Bai, Lifeng Shang} \\
  Huawei Technologies}

\hypersetup{
  pdftitle={HARPO: Hallucination-Aware Reinforcement Learning for Faithful and Creative Language Generation},
  pdfauthor={Tiezheng Yu, Yuxin Jiang, Jinpeng Li, Shuning Sun, Fei Mi, Haoli Bai, Lifeng Shang}
}

\begin{document}
\maketitle
% !TeX root = ../acl_latex.tex
\begin{abstract}
% Large Language Models (LLMs) often generate hallucinated content that undermines reliability, especially in faithful tasks. To address this without sacrificing creativity, we propose \textbf{HARPO} (Hallucination-Aware Reinforcement for Policy Optimization), a reinforcement learning framework that jointly optimizes faithfulness and creative quality. HARPO employs a Generative Reward Model (GRM) trained via verifiable rewards to assess both hallucination and writing quality, combined with a Selective Activation Mechanism (SAM) that activates writing rewards only when outputs are hallucination-free. A curriculum learning scheme further stabilizes training by gradually shifting from general to hallucination-focused data. Experiments show that HARPO significantly improves faithfulness while preserving expressive generation, outperforming strong open-source and proprietary baselines.

Large Language Models (LLMs) are prone to generating hallucinated content, which compromises their reliability in knowledge-intensive tasks. To address this challenge without sacrificing creativity, we propose \textbf{HARPO}, a reinforcement learning framework designed to jointly optimize faithfulness and creativity. HARPO incorporates a Hallucination-Aware Generative Reward Model (HA-GRM), trained via verifiable feedback, to assess both faithfulness and writing quality. A Selective Activation Mechanism (SAM) activates writing rewards only for outputs judged hallucination-free by HA-GRM, while a data curriculum progressively shifts training from creative writing to hallucination-centric tasks. On RAGTruth, our Qwen3-4B-based HA-GRM achieves a response-level F1 score of 78.08\%, compared with 66.37\% for the supervised fine-tuning baseline. Experiments on Qwen2.5 and Qwen3 models from 1.7B to 8B parameters show improvements in both faithful generation and writing quality. On Qwen3-4B, HARPO reduces the HA-GRM-judged hallucination rate on MultiHopRAG from 3.29\% to 1.02\%, while increasing the Arena-Hard-v2.0 creative-writing score from 16.95\% to 27.54\%.

\end{abstract}

% !TeX root = ../acl_latex.tex
\section{Introduction}
Large Language Models (LLMs) have demonstrated impressive capabilities in generating fluent, human-like text. However, their reliability remains a critical concern due to hallucinated outputs that are unfaithful with respect to a given source~\cite{Ji2022SurveyOH}. Such inaccuracies undermine the trustworthiness and utility of LLMs, particularly in high-stakes applications.

% Recently, reinforcement learning (RL) techniques such as Group Relative Policy Optimization (GRPO) have been widely adopted in the post-training phase to improve LLM performance~\cite{wei2025truthrl}. Yet, several works have shown that optimizing a single capability in isolation can degrade performance in other dimensions. For example, RLMR~\cite{liao2025rlmr} demonstrated the difficulty of simultaneously maintaining high subjective writing quality and strict adherence to objective constraints. Similarly, RLCR~\cite{damani2025beyond} reported that after RL training on HotpotQA~\cite{yang2018hotpotqa}, the resulting model became substantially less reliable at assessing its own confidence on out-of-domain benchmarks such as MATH-500~\cite{lightman2023let}.
While reinforcement learning (RL) techniques have been widely adopted to boost the performance of large language models (LLMs)~\cite{wei2025truthrl}, optimizing for a single capability often comes at the cost of degraded performance in other dimensions. For instance, the RLMR framework~\cite{liao2025rlmr} underscores the inherent challenge of preserving subjective textual quality while adhering to rigid objective constraints. Similarly, \citet{damani2025beyond} show that conventional RL can impair calibration on out-of-domain benchmarks and propose RLCR to jointly optimize correctness and calibration.

% In this paper, we focus on reducing hallucinations while preserving creative writing capabilities. Mitigating hallucinations requires the model to produce precise and cautious responses, avoiding content that cannot be directly inferred from the given context. In contrast, creative generation tasks demand richer linguistic expression and greater novelty in content. Existing methods struggle to balance these competing objectives: single-reward strategies fail to improve both abilities simultaneously, and simply mixing training data tends to cause interference between optimization along the two dimensions of hallucination and creative writing.

In this paper, we address the conflict between \textbf{hallucination mitigation} and \textbf{creative generation}. Mitigating hallucinations demands precision and caution, whereas creative writing requires richness and novelty. Existing methods struggle to balance these competing objectives: single-reward strategies fail to improve both simultaneously, while naive data mixing often leads to interference between faithful adherence and expressive quality.

To address this problem, we introduce \textbf{HARPO} (\underline{H}allucination-\underline{A}ware \underline{R}einforcement for \underline{P}olicy \underline{O}ptimization), a framework designed to jointly optimize faithfulness and creativity. Prior work has established reinforcement learning with verifiable span-level rewards for hallucination detection~\citep{su2025learning}. HARPO combines hallucination evaluation and writing preference evaluation in a Hallucination-Aware Generative Reward Model (HA-GRM), trained using Reinforcement Learning with Verifiable Rewards (RLVR)~\citep{shao2024deepseekmath}. We combine response-level classification with an auxiliary reward based on hallucinated-span prediction to supervise faithfulness evaluation. By incorporating pairwise preference data for creative writing, HA-GRM also learns to evaluate stylistic quality.

To effectively utilize these signals during policy optimization, we propose a Selective Activation Mechanism (SAM). Unlike traditional fixed-weight scalarization, prone to reward conflicts, SAM acts as a conditional gate: the writing reward is activated only if HA-GRM judges the response to be hallucination-free. Finally, to stabilize the training process, we design a data curriculum strategy that gradually shifts the data distribution from creative writing instructions to hallucination-focused scenarios. This approach mitigates the catastrophic forgetting often observed in single-objective alignment, preserving broad linguistic proficiency while sharpening evidential rigor.

% We validate the effectiveness of HARPO through extensive experiments on the Qwen2.5 and Qwen3 model families across scales ranging from 1.7B to 8B parameters. Our results demonstrate that HA-GRM achieves superior generalization on unseen hallucination tasks compared to standard supervised fine-tuning, largely due to the dense supervision provided by span-level rewards. When applied to policy optimization, HARPO consistently delivers Pareto improvements, significantly reducing hallucination rates, by over 69\% in the case of Qwen2.5-7B, while simultaneously boosting creative writing performance. Crucially, comparative analyses reveal that our Selective Activation Mechanism outperforms traditional linear reward scalarization by preventing reward hacking, while our curriculum strategy successfully mitigates the catastrophic forgetting often observed in single-objective alignment.
% We validate HARPO through extensive experiments on the Qwen2.5 and Qwen3 model families (1.7B to 8B parameters). 
HA-GRM achieves higher Overall response-level F1 than the base model and SFT, while SFT remains stronger on QA detection and the base model retains a higher D2T F1. We evaluate HARPO on Qwen2.5 and Qwen3 models from 1.7B to 8B parameters. Relative to the base Qwen3-4B model, HARPO reduces the HA-GRM-judged summarization hallucination rate by over 85\% while increasing creative-writing scores. Compared with linear reward mixing on Qwen3-4B, HARPO has a lower summarization hallucination rate and a higher General score, with a slightly higher MultiHopRAG hallucination rate and a lower creative-writing score. The curriculum comparison similarly shows a trade-off between hallucination rates and writing performance.

The contributions are as follows:
\begin{itemize}
    \item We introduce HARPO, a framework that combines hallucination-aware reward modeling, conditional reward aggregation, and a creative-to-faithful data curriculum to jointly optimize faithful generation and writing quality.
    \item We train HA-GRM with response-level labels, span-level annotations, and writing preferences, and use SAM to activate writing rewards for responses predicted to be hallucination-free by HA-GRM.
    \item Extensive experiments demonstrate that HARPO consistently enhances performance across various model families and scales, confirming its effectiveness.
\end{itemize}

\section{Methodology}
\begin{figure*}[!ht]
    \centering 
    \includegraphics[width=1\textwidth]{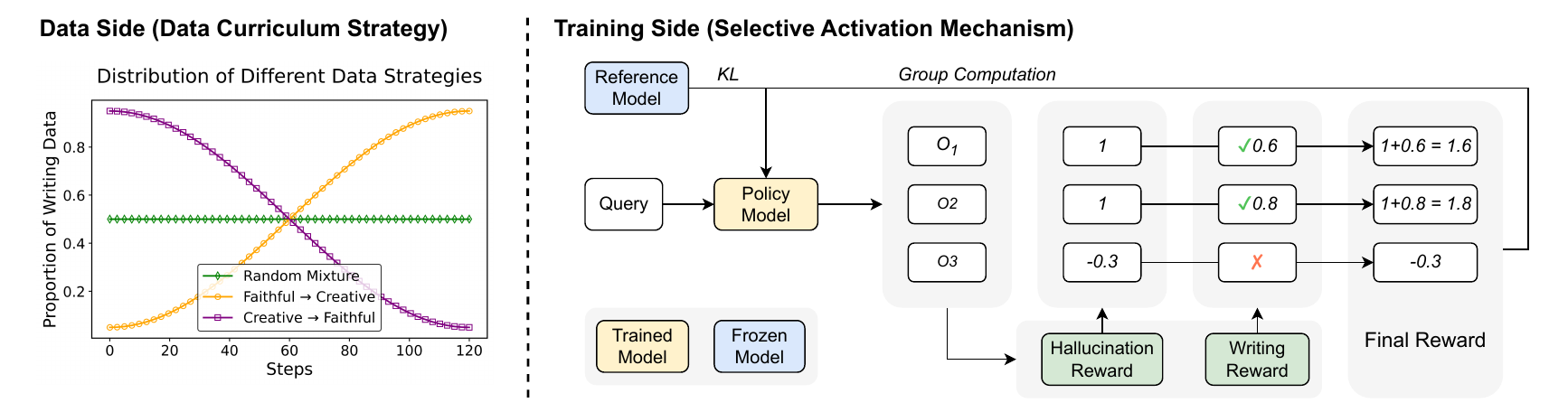} 
    \caption{\textbf{The HARPO Framework.} Left: The \textbf{Data Curriculum Strategy} smoothly shifts the training distribution from creative to faithful tasks via cosine annealing. Right: The \textbf{Selective Activation Mechanism} activates writing rewards only for responses judged hallucination-free by HA-GRM.}
    \label{fig:Framework} 
\end{figure*}
We introduce HARPO, a framework designed to mitigate hallucinations in large language models (LLMs) while preserving creative generation capabilities. Our approach integrates a Hallucination-Aware Generative Reward Model (HA-GRM) with a curriculum-driven reinforcement learning strategy. The optimization process features two core mechanisms: a Selective Activation Mechanism (SAM) that gates writing rewards based on HA-GRM's hallucination predictions, and a dynamic curriculum that progressively shifts the training distribution from creative to hallucination-sensitive tasks.

\subsection{HA-GRM: Reward Model Training}
\label{sec:HA-GRM}
% \paragraph{Task Formulation.} In terms of task formulation, generative reward models and hallucination detectors follow the same paradigm. Following~\citep{Wu2023RAGTruthAH}, we train the Hallucination-Aware GRM to detect both response-level and span-level hallucinations. The problem is formally defined as follows: Given the input context $c$ and the generated response $y = (y_{1}, y_{2}, \ldots, y_{T})$ consisting of $T$ characters, the model needs to first identify whether the response $y$ is hallucinated and then locate all hallucinated spans $S$, defined as portions of text in $y$ that lack support from $c$. Besides, we jointly incorporate pairwise preference data for improving the creative writing judgment ability. Similar to the Arena-Hard~\citep{li2024crowdsourced}, We encourage the Reward Model to evaluate the responses from five perspectives, namely: helpfulness, relevance, conciseness, creativity and completeness.
\paragraph{Task Formulation.} We train the HA-GRM to detect hallucinations at both the response and span levels, following the annotation structure of RAGTruth~\citep{Wu2023RAGTruthAH}. Given the input context $c$ and the generated response $y = (y_{1}, y_{2}, \ldots, y_{T})$ consisting of $T$ characters, the model needs to first identify whether the response $y$ is hallucinated and then locate all hallucinated spans $S$, defined as portions of text in $y$ that lack support from $c$. Additionally, to maintain expressive quality, we incorporate pairwise preference data, encouraging the model to evaluate responses based on helpfulness, relevance, conciseness, creativity, and completeness, similar to Arena-Hard~\citep{li2024crowdsourced}.

\paragraph{Rule-Based Reinforcement Learning.} We fine-tune the HA-GRM using rule-based online RL, adopting the GRPO setting~\citep{shao2024deepseekmath}. During rollout, the model predicts hallucinated spans and pointwise response-level labels for hallucination detection, while generating pairwise comparisons for creative writing. The reward signal combines format validity and prediction accuracy. 

For hallucination detection, we incorporate the span-F1 metric alongside response-level accuracy. Let $\hat{S}$ be the predicted hallucination spans and $S$ be the ground-truth spans. The hallucination reward $r_h$ is

\begin{equation}
r_{h} =
\begin{cases}
1 + \text{span-F1}(S, \hat{S}), & \text{valid format \& acc.} \\
1, & \text{valid format only} \\
0, & \text{invalid format}
\end{cases}
\label{eq:rating_rule_hallucination}
\end{equation}

For writing preference evaluation, the reward $r_{w}$ is

\begin{equation}
r_{w} = 
\begin{cases}
2, & \text{correct format \& acc.} \\
1, & \text{correct format only} \\
0, & \text{incorrect format}
\end{cases}
\label{eq:rating_rule_writing}
\end{equation}

These reward schemes are applied separately to the pointwise hallucination data and pairwise creative writing data.

\subsection{HARPO: Hallucination-Aware Reinforcement for Policy Optimization}
We now present the details of HARPO, a framework designed to balance the competing objectives of hallucination reduction and creative generation. We employ reinforcement learning rather than direct Supervised Fine-Tuning. This approach enables us to optimize the model using unlabelled prompts without requiring ground-truth reference texts. To guide this process, we introduce two novel components: first, the Selective Activation Mechanism, which functions as the core scoring metric during optimization; and second, a data curriculum strategy that schedules the training distribution to smooth the transition between hallucination-centric and creativity-centric tasks.

\paragraph{Reinforcement Learning with GRPO.}
As our reinforcement learning framework, we adopt Group Relative Policy Optimization (GRPO)~\citep{shao2024deepseekmath}, a variant of REINFORCE~\citep{williams1992simple} that utilizes group-based advantage normalization. To optimize the LLM policy $\pi_\theta$, given a prompt $x$, we sample a group of $K$ rollouts $\{\tau_k\}^K_{k=1}$ from the current policy $\pi_\theta$. Each rollout $\tau_k$ is assigned a reward score $R_k = R(\tau_k)$ calculated by our reward system. The policy is updated via the clipped policy gradient objective:

\begin{align}
\mathcal{L}_{\text{GRPO}} &= \frac{1}{K} \sum_{k=1}^{K} \frac{1}{|\tau_k|} \sum_{t=1}^{|\tau_k|} \min \left[ w_{k,t}(\theta) \cdot \hat{A}_{k,t}, \right. \notag\\
& \quad \left. \text{clip}(w_{k,t}(\theta), 1-\epsilon_{\text{low}}, 1+\epsilon_{\text{high}}) \cdot \hat{A}_{k,t} \right] \notag\\
& \quad - \beta \cdot D_{\text{KL}}(\pi_\theta \parallel \pi_{\text{ref}}).
\end{align}

\noindent where the token-level importance weight is defined as the probability ratio:
\begin{align}
w_{k,t}(\theta) = \frac{\pi_\theta(\tau_{k,t} | x, \tau_{k,<t})}{\pi_{\theta_{\text{old}}}(\tau_{k,t} | x, \tau_{k,<t})}.
\end{align}
Here, $\tau_{k,t}$ denotes the token generated at step $t$ in rollout $k$, and $\tau_{k,<t}$ denotes the preceding context. The group-normalized advantage $\hat{A}_{k,t}$ is computed using the rewards of the $K$ rollouts sampled for the same input:
\begin{align}
\hat{A}_{k,t} = \frac{R(\tau_k) - \text{mean}(\{R_i\}^K_{i=1})}{\text{std}(\{R_i\}^K_{i=1}) + \delta}.
\end{align}
where $\delta$ is a small constant for numerical stability. Following recent advances in RL for LLM reasoning such as DAPO~\citep{yu2025dapo}, we adopt asymmetric clipping (clip-higher) with $\epsilon_{\text{low}} < \epsilon_{\text{high}}$. This strategy encourages the exploration of novel token sequences while maintaining training stability. The coefficient $\beta$ scales the KL divergence term $D_{\text{KL}}(\pi_\theta \parallel \pi_{\text{ref}})$, ensuring the policy remains close to the reference model $\pi_{\text{ref}}$ to prevent mode collapse.

\paragraph{Selective Activation Mechanism (SAM).}
A core challenge in aligning LLMs for both reliability and quality is the potential conflict between objectives. Standard approaches typically employ a linear scalarization of rewards, $R_{\text{total}} = w_1 R_{\text{hallu}} + w_2 R_{\text{creat}}$. However, this fixed-weight strategy creates an optimization landscape where the model may learn to exploit the reward function. We propose the Selective Activation Mechanism (SAM), a hierarchical reward aggregation strategy that activates the writing reward only when HA-GRM predicts no hallucinated spans.
% For example, by generating highly creative but hallucinated content to maximize the weighted sum, effectively bypassing the factuality constraint.

Let $x$ denote the input prompt and $y$ denote the generated response corresponding to a rollout $\tau_k$. Our HA-GRM, described in Section~\ref{sec:HA-GRM}, provides two distinct scalar signals:

\begin{enumerate}
    \item \textbf{Hallucination Reward ($R_{\text{hallu}}$):} A score based on HA-GRM's prediction of whether the response $y$ is grounded in the context. Let $\hat{S}$ denote the set of hallucinated spans predicted by HA-GRM, and let $|\cdot|$ denote the text length. The hallucination reward is calculated as:
    \begin{equation}
        R_{\text{hallu}}(x, y) = 
        \begin{cases} 
        1, & \text{if } \hat{S} = \emptyset  \\
        -\frac{|\hat{S}|}{|y|}, & \text{if } \hat{S} \neq \emptyset 
        \end{cases}
    \end{equation}
    This formulation imposes a penalty proportional to the density of hallucinated content when errors occur.

    \item \textbf{Writing Reward ($R_{\text{creat}}$):} A score evaluating the helpfulness, relevance, conciseness, and novelty of the response. This score is derived from a pairwise evaluation comparing the model's output $y$ against a reference answer $r$.
\end{enumerate}

Under the SAM framework, the writing reward is gated by HA-GRM's hallucination prediction. The final reward $R_{\text{SAM}}$ assigned to a rollout is defined as:
\begin{align}
R_{\text{SAM}}(x, y, r) &= R_{\text{hallu}}(x, y) \\
& \quad + \mathbb{I}[\hat{S} = \emptyset] \cdot R_{\text{creat}}(x, y, r). \notag
\end{align}
where $\mathbb{I}[\cdot]$ is the indicator function and $\hat{S}$ is the set of hallucinated spans predicted by HA-GRM. The writing reward is activated when $\hat{S} = \emptyset$ and suppressed otherwise. This gate constrains reward assignment according to HA-GRM's prediction; it does not guarantee that the response is truly hallucination-free.

\paragraph{Data Curriculum Strategy.}
The distinct nature of our two types of training data, creative writing and hallucination mitigation, presents a significant optimization challenge. Creative writing tasks demand open-ended generation with diverse vocabulary, whereas hallucination mitigation requires strict adherence to evidence and factual constraints. Directly mixing these datasets in a static ratio often leads to optimization interference.

To address this, we design a dynamic data curriculum that modulates the training distribution over time. We adopt a \textit{Creative $\to$ Faithful} progression strategy for two reasons: (1) \textbf{Model Affinity:} We initialize training with a higher proportion of creative writing data. This leverages the pre-trained model's inherent linguistic capabilities, stabilizing the initial policy optimization before introducing strict factual constraints. (2) \textbf{Prevention of Catastrophic Forgetting:} Rather than abruptly switching tasks, we employ a smooth decay schedule. This ensures that even as the focus shifts toward hallucination mitigation, a minimum proportion of creative writing data is maintained to preserve language proficiency.

Formally, let $T$ denote the total number of training steps. We define $r(t)$ as the sampling ratio of creative writing data at step $t$. We utilize a cosine annealing schedule to smoothly reduce this ratio from an initial high to a fixed minimum:

\begin{align}
r(t) = r_0 \cdot \cos \left(\frac{\pi}{2T} \cdot t\right) + r_{\min}.
\end{align}

\noindent where $r_{\min} = 0.05$ represents the minimum proportion of creative data maintained at the end of training to prevent forgetting, and $r_0 = 0.9$ represents the magnitude of the decay. Under this schedule, the training begins with a creative data ratio of $r(0) = r_0 + r_{\min}$ and gradually converges to $r(T) = r_{\min}$ as the model becomes more robust to hallucination-sensitive tasks.

% !TeX root = ../acl_latex.tex
\section{Experiments on HA-GRM}
\label{sec:epx-HA-GRM}

\begin{table*}[!ht]
\centering
\small
\setlength{\tabcolsep}{3pt}
\begin{tabular}{l|ccc@{\hspace{8pt}}ccc@{\hspace{8pt}}ccc|ccc}
\toprule
\multirow{2}{*}{Model Name} & \multicolumn{3}{c}{\textsc{RAGTruth (Summ)}} & \multicolumn{3}{c}{\textsc{RAGTruth (QA)}} & \multicolumn{3}{c}{\textsc{RAGTruth (D2T)}} & \multicolumn{3}{c}{\textsc{Overall}}\\
\cmidrule(lr){2-4} \cmidrule(lr){5-7} \cmidrule(lr){8-10}  \cmidrule(lr){11-13}
&P &R & F1 &P &R & F1 &P &R & F1 &P &R & F1 \\

\midrule
\rowcolor{gray!10} \multicolumn{13}{l}{\textit{Open-Source Models (8B - 671B)}} \\
Qwen3-8B                        &50.14 &87.25 &63.69    &36.81 &88.12 &51.93   &79.76 &91.19 &85.09   &60.50 &89.82 &72.30  \\
Qwen3-32B                       &42.76 &91.18 &58.22    &33.64 &91.88 &49.25   &80.12 &94.28 &\textbf{86.62}   &56.54 &93.20 &70.39  \\
Qwen3-235B-A22B-Instruct        &40.25 &96.55 &56.81    &28.65 &95.62 &44.09   &66.55 &99.48 &79.75   &48.99 &98.19 &65.37  \\
Qwen3-235B-A22B-Thinking        &37.94 &97.01 &54.55    &28.95 &95.00 &44.38   &71.84 &98.44 &83.07   &50.03 &97.55 &66.14  \\
DeepSeek-V3.2                   &32.05 &96.97 &48.18    &26.83 &96.86 &42.02   &65.75 &99.83 &79.28   &44.97 &98.71 &61.79  \\

\midrule
\rowcolor{gray!10} \multicolumn{13}{l}{\textit{Proprietary Models}} \\
GPT-4o                          &50.29 &86.43 &63.59   &32.48 &86.39 &47.21   &69.60 &96.98 &81.04   &55.67 &92.95 &69.64  \\
doubao-seed-1-6-thinking        &41.81 &95.10 &58.08   &37.29 &96.25 &53.75   &74.47 &98.26 &84.73   &55.83 &97.23 &70.94  \\
claude-4-sonnet                 &46.32 &86.27 &60.27   &36.65 &87.50 &51.66   &72.95 &94.91 &82.49   &54.83 &91.21 &68.49  \\

\midrule
\rowcolor{gray!10} \multicolumn{13}{l}{\textit{Open-Source Models (Qwen3-4B based)}} \\
Qwen3-4B                        &60.92 &71.43 &65.76    &43.66 &73.12 &54.67    &84.75 &86.09 &85.41    &60.92 &80.67 &74.58 \\
HA-GRM (\textit{SFT})    &78.91 &56.86 &66.10    &71.61 &69.38 &\textbf{70.48}    &81.61 &52.33 &65.09    &81.04 &56.20 &66.37 \\
HA-GRM (\textit{Response Only})  &77.44 &62.25 &69.02    &63.79 &69.38 &66.47    &91.22 &77.34 &83.71    &82.73 &72.72 &77.40 \\
HA-GRM (\textit{Full})           &81.05 &60.78 &\textbf{69.47}    &70.75 &65.00 &67.75    &91.98 &77.20 &83.94    &85.88 &71.58 &\textbf{78.08}  \\

\bottomrule
\end{tabular}
\caption{Performance (Precision, Recall and F1) of hallucination evaluation on the test split of RAGTruth. We separate Qwen3-4B based models, larger open-source models, and proprietary models into different blocks.}
\label{tab:grm_response_level_results}
\end{table*}
We address the following research question: (\textbf{RQ1}) What are the advantages of using reinforcement learning with fine-grained rewards for training generative reward models?
\subsection{Experimental Setup}
\paragraph{Training Details.} We initialize HA-GRM from Qwen3-4B and train it on two public datasets. Specifically, for hallucination detection, we use the summarization and question-answering subsets of the \textsc{RAGTruth}~\cite{Wu2023RAGTruthAH} training set. For creative writing, we utilize \textsc{Arena-Hard-v2.0}~\citep{li2024crowdsourced}. We employ verl~\cite{sheng2025hybridflow} as our RL training framework, using a global batch size of 128 and a learning rate of 1e-6. Additionally, we apply a clipping strategy with the upper clipping threshold set to 0.28.

\paragraph{Evaluation Datasets and Metrics.} We evaluate the reward models on the RAGTruth test set for hallucination detection and the Auto-J pairwise preference evaluation data~\cite{ligenerative} for general preference evaluation. In addition to response-level metrics such as precision, recall, and F1 scores, we conduct span-level detection by calculating the overlap between detected and human-labeled spans, reporting the character-level F1 score. For the pairwise evaluation on Auto-J, we report accuracy. All datasets used in this evaluation are human-annotated.

\subsection{Results and Analysis}
\paragraph{Main Results.}
Table~\ref{tab:grm_response_level_results} reports response-level detection performance. HA-GRM (\textit{Full}), initialized from Qwen3-4B, achieves an Overall F1 of 78.08\%, compared with 74.58\% for the base model and 66.37\% for SFT. The gains vary by task: Full has a higher summarization F1 than SFT (69.47\% vs. 66.10\%), while SFT has a higher QA F1 (70.48\% vs. 67.75\%). On D2T, which is excluded from our training data, Full reaches 83.94\%, above SFT's 65.09\% but below the base model's 85.41\%. Full also has the highest reported Overall F1 among the compared models, while Qwen3-32B retains the highest D2T F1 (86.62\%). These results support improved overall detection performance with task-specific trade-offs.

\paragraph{Can span-level hallucination prediction help?}
\begin{table}[!ht]
\centering
\small
\setlength{\tabcolsep}{3pt}
\resizebox{\linewidth}{!}{
\begin{tabular}{l|cccc}
\toprule
Model Name & \textsc{Summ} & \textsc{QA} & \textsc{D2T} & \textsc{Overall}\\
\midrule
Qwen3-4B                        &41.43 &34.00 &38.08 &37.33  \\
HA-GRM (\textit{SFT})    &45.94 &\textbf{61.26} &30.13 &44.58 \\
HA-GRM (\textit{Response Only}) &48.53 &48.74 &39.31 &44.43 \\
HA-GRM (\textit{Full})           &\textbf{52.10} &53.63 &\textbf{47.26} &\textbf{50.38} \\
\bottomrule
\end{tabular}
}
\caption{Span-level hallucination detection performance (span-F1) on the RAGTruth test set. Bold indicates the best score in each column.}
\label{tab:grm_span_level_results}
\end{table}
% 这里分析 span-level prediction 的优势以及能帮助提升response level prediction.
To investigate span supervision, we compare HA-GRM \textit{(Full)} with \textit{(Response Only)}, which is trained solely with response-level binary rewards. In Table~\ref{tab:grm_span_level_results}, Full has higher span-F1 on each subset than Response Only, with Overall span-F1 increasing from 44.43\% to 50.38\%. However, SFT retains the highest QA span-F1 (61.26\%, compared with Full's 53.63\%). Table~\ref{tab:grm_response_level_results} also shows a response-level Overall F1 increase from 77.40\% to 78.08\% for Full over Response Only. Thus, span supervision improves the response-only variant on the reported metrics, while SFT remains stronger on QA span detection.

\paragraph{How does writing preference evaluation affect hallucination detection?}
\begin{table}[!ht]
\centering
\small
\setlength{\tabcolsep}{2.5pt}
\resizebox{\linewidth}{!}{
\begin{tabular}{l|ccc}
\toprule
Model Name & \textsc{RAGTruth} & \textsc{Auto-J} & \textsc{Average}\\
\midrule
Qwen3-4B                                               &74.58 &57.22 &65.90  \\
\hspace{3mm} \textit{w/ Hallucination}                 &\textbf{78.41} &57.31 &67.86   \\
\hspace{3mm} \textit{w/ Hallucination and Writing} &78.08 &\textbf{59.91} &\textbf{69.00} \\
\bottomrule
\end{tabular}
}
\caption{Impact of multi-objective training on hallucination detection (RAGTruth) and general preference evaluation (Auto-J) test sets.}
\label{tab:grm_hallu_writing_results}
\end{table}

Table~\ref{tab:grm_hallu_writing_results} shows the trade-off from adding writing preference data. Compared with hallucination-only training, joint training increases Auto-J accuracy from 57.31\% to 59.91\%, while RAGTruth response-level F1 decreases from 78.41\% to 78.08\%. The reported cross-task average rises from 67.86\% to 69.00\%. Joint training therefore improves preference evaluation with a small reduction in hallucination detection F1; the two task metrics should be considered separately when choosing a training configuration.
\begin{takeaway}
    {Takeaway (RQ1)}
    {
        HA-GRM improves Overall response-level F1 over the base model and SFT, with task-specific trade-offs. Adding writing preference data raises Auto-J accuracy with a small decrease in RAGTruth F1.
    }
\end{takeaway}

% This result confirms that our multi-objective alignment strategy successfully decouples factuality from creativity, allowing the reward model to appreciate stylistic quality without compromising its rigor in detecting factual errors.

% !TeX root = ../acl_latex.tex
\section{Experiments on HARPO}
We aim to answer the research question: (RQ2) Can we leverage the feedback from HA-GRM to effectively mitigate LLM hallucinations while incentivizing its creative writing capabilities?

\subsection{Experimental Setup}
\paragraph{Datasets and Evaluation Metrics.} Our training corpus consists of three distinct datasets selected to balance faithful grounding with creative expression. For hallucination mitigation, we utilize the summarization subset of \textsc{RAGTruth}~\citep{Wu2023RAGTruthAH} and the question-answering subset of \textsc{HaluEval}~\citep{li2023halueval}. To support creative generation, we curate a subset of the \textsc{Arena-Human-Preference-140k} dataset from Chatbot Arena~\citep{chiang2024chatbot},\footnote{\url{https://huggingface.co/datasets/lmarena-ai/arena-human-preference-140k}} filtering specifically for prompts related to creative writing. All experiments are run once.

For summarization, we evaluate on 831 source documents from the Vectara Hallucination Leaderboard dataset~\citep{hughes2023vectara}, denoted \textsc{HHEM} in our tables. For QA, we use \textsc{MultiHopRAG}~\citep{tang2024multihoprag}. Both hallucination rates are the percentage of generated responses judged hallucinated by our best-performing HA-GRM. General and creative writing capabilities are evaluated on the Hard Prompts and Creative Writing subsets of \textsc{Arena-Hard-v2.0}, respectively, using its official pipeline~\citep{li2024crowdsourced}.\footnote{\url{https://github.com/lmarena/arena-hard-auto}} Table~\ref{tab:harpo_dataset} summarizes the statistics for all datasets.

\begin{table}[!ht]
\centering
\small
\setlength{\tabcolsep}{3pt}
\resizebox{\linewidth}{!}{
\begin{tabular}{l|lc}
\toprule
Dataset & \textsc{Domain} & \textsc{\# of Samples} \\
\midrule
\rowcolor{gray!10} \multicolumn{3}{l}{\textit{Training data}} \\
RAGTruth               &Summarization &2,217   \\
HaluEval &Question Answering &1,452 \\
Arena-Human-Preference-140k &Creative Writing &3,000 \\
\midrule
\rowcolor{gray!10} \multicolumn{3}{l}{\textit{Testing data}} \\
HHEM &Summarization &831   \\
MultiHopRAG &Question Answering &2,556 \\
Arena-Hard-v2.0 (Hard Prompts) &General &500 \\
Arena-Hard-v2.0 (Creative Writing) &Creative Writing &250 \\
\bottomrule
\end{tabular}
}
\caption{Statistics of the training and testing datasets utilized in the HARPO experiments.}
\label{tab:harpo_dataset}
\end{table}

\paragraph{Baselines and Training Details.} To evaluate HARPO, we compare it against three baselines: (1) \textbf{\textit{Hallucination-Only}}, trained exclusively on \textsc{RAGTruth} and \textsc{HaluEval} using only $R_{\text{hallu}}$; (2) \textbf{\textit{Writing-Only}}, trained solely on \textsc{Arena-Human-Preference-140k} using only $R_{\text{creat}}$; and (3) \textbf{\textit{Linear Mixture}}, a combined-objective baseline. Following~\citet{peng2025agentic}, the Linear Mixture baseline performs multi-objective alignment via fixed scalarization. We normalize the writing reward to $[0,1]$ and compute the arithmetic mean with the hallucination reward: $(R^{\text{norm}}_{\text{creat}} + R_{\text{hallu}}) / 2$. This serves as the primary state-of-the-art comparison for mixed-objective training. This baseline uses the same training data as HARPO but a different reward strategy. For reinforcement learning hyperparameters, we adopt the same configuration used to train HA-GRM.
 
\begin{table*}[!ht]
\centering
\small
\setlength{\tabcolsep}{5pt}
\begin{tabular}{l|cccc}
\toprule
\multirow{2}{*}{Model Name} & \multicolumn{2}{c}{Hallucination Control $\downarrow$} & Creative Writing $\uparrow$  & General $\uparrow$ \\
\cmidrule(lr){2-3} \cmidrule(lr){4-4} \cmidrule(lr){5-5}
& \textsc{HHEM} & \textsc{MultiHopRAG}    & \makecell[c]{\textsc{Arena-Hard-v2.0}\\(Creative Writing)} & \makecell[c]{\textsc{Arena-Hard-v2.0}\\(Hard Prompts)}\\
\midrule
\rowcolor{gray!10} \multicolumn{5}{l}{\textit{Qwen2.5 series}}                \\
Qwen2.5-3B                              &11.07 &4.69 &2.33  &2.07             \\
\hspace{3mm} \textit{w/ HARPO (Ours)}    &2.41  &2.86 &2.66  &2.30             \\
\midrule
Qwen2.5-7B                              &5.05  &1.88 &4.68  &4.63             \\
\hspace{3mm} \textit{w/ HARPO (Ours)}    &1.56  &0.86 &6.78  &4.93             \\

\midrule
\rowcolor{gray!10} \multicolumn{5}{l}{\textit{Qwen3 series}}                  \\
Qwen3-1.7B                                     &11.07 &5.38 &7.04   &3.87     \\
\hspace{3mm} \textit{w/ HARPO (Ours)}           &3.97  &2.03 &8.53   &4.05     \\
\midrule
Qwen3-4B                                       &7.34  &3.29 &16.95  &11.16    \\
\hspace{3mm} \textit{w/ Hallucination-Only}    &0.72  &0.98 &9.74   &5.29     \\
\hspace{3mm} \textit{w/ Writing-Only}       &24.31 &2.66 &43.35  &11.45    \\
\hspace{3mm} \textit{w/ Linear Mixture}        &3.73  &0.90 &28.83  &5.63     \\
\hspace{3mm} \textit{w/ HARPO (Ours)}          &1.08  &1.02 &27.54  &12.35    \\
\midrule
Qwen3-8B                                       &4.93  &2.90 &33.76  &16.08    \\
\hspace{3mm} \textit{w/ HARPO (Ours)}           &1.56  &1.17 &41.43  &16.52    \\
\bottomrule
\end{tabular}
\caption{Comparison of HARPO against base models and baselines. ($\downarrow$) indicates lower is better, ($\uparrow$) indicates higher is better. HARPO lowers HA-GRM-judged hallucination rates and improves writing scores relative to the corresponding base models.}
\label{tab:main_results}
\end{table*}

\subsection{Results and Analysis}
\paragraph{Main Results.}
% In this part, I want to analyze these three parts:
% 1) Analyze main results and different series and size of models. 
% 2) Show that only train on hallucination or creativity data will effect the other domain performance. 
% 3) The advantage of our Selective Activation Mechanism because compare with Linear Mixture baseline.
% 4) others, how hallucination control, creative writing and general affect each other.
Table~\ref{tab:main_results} presents the performance of HARPO across different model families (Qwen2.5, Qwen3) and parameter scales (1.7B to 8B). Our analysis yields two key observations:

\noindent \textbf{1. Trade-offs in Single-Objective Training:}
On Qwen3-4B, \textit{Hallucination-Only} reaches the lowest HHEM hallucination rate (0.72\%), while Creative Writing decreases from 16.95\% to 9.74\% and General decreases from 11.16\% to 5.29\% relative to the base model. \textit{Writing-Only} raises Creative Writing from 16.95\% to 43.35\%, while the HHEM hallucination rate rises from 7.34\% to 24.31\%. These results show that optimizing either objective alone can improve its target metric while degrading another.

\noindent \textbf{2. Improvements over Base Models:}
HARPO has lower HA-GRM-judged hallucination rates on both datasets and higher Creative Writing and General scores than the corresponding base models at every tested size. For example, on Qwen2.5-7B, the HHEM hallucination rate decreases from 5.05\% to 1.56\% and Creative Writing increases from 4.68\% to 6.78\%. These comparisons concern the base models; comparisons with other training strategies reveal additional trade-offs.

\paragraph{Comparison with Linear Reward Mixing.}
For Qwen3-4B, HARPO has a lower HHEM hallucination rate than \textit{Linear Mixture} (1.08\% vs. 3.73\%) and a higher General score (12.35\% vs. 5.63\%). However, its MultiHopRAG hallucination rate is slightly higher (1.02\% vs. 0.90\%), and its Creative Writing score is slightly lower (27.54\% vs. 28.83\%). HARPO therefore improves summarization faithfulness as judged by HA-GRM and performance on Hard Prompts, with a trade-off in QA hallucination rate and creative-writing score. SAM activates writing rewards only for responses judged hallucination-free by HA-GRM; this comparison reflects the complete HARPO training configuration.

\paragraph{Effect of Data Curriculum Strategy.}

%  In this paragraph, I want to analyze the advantage of our Data Curriculum Strategy. Random means random mix two types of data, Progressive Writing means the beginning point is hallucination dominant and progressively move to writing dominant. The Progressive Hallucination is our proposed Data Curriculum Strategy.

\begin{table*}[t]
\centering
\small
\setlength{\tabcolsep}{5pt}
\begin{tabular}{l|ccc}
\toprule
\multirow{2}{*}{Curriculum Schedule} & \multicolumn{2}{c}{Hallucination Control $\downarrow$} & Creative Writing $\uparrow$ \\
\cmidrule(lr){2-3} \cmidrule(lr){4-4}
& \textsc{HHEM} & \textsc{MultiHopRAG}    & \makecell[c]{\textsc{Arena-Hard-v2.0}\\(Creative Writing)}\\
\midrule
Qwen3-4B                                          &7.34 &3.29 &16.95 \\
Random Mixture (Static)                           & 4.09 & 1.17 & 23.65                   \\
Faithful $\to$ Creative (Reverse)                 & 9.64 & 1.72 & \textbf{34.30}          \\
Creative $\to$ Faithful (\textbf{Ours})           & \textbf{1.08} & \textbf{1.02} & 27.54 \\

\bottomrule
\end{tabular}
\caption{Ablation on Data Curriculum Strategies.}
\label{tab:curriculum_learning_strategy}
\end{table*}
% Here we compare three different data mixing strategy, 1) Randomly mix data from two tasks 2) Curriculum from writing to hallucination, 3) curriculum from hallucination to writing. 
To validate our data curriculum strategy, we compare it against two alternative schedules as shown on the left of Figure~\ref{fig:Framework}: (1) \textit{Random Mixture}, which samples tasks with a fixed probability throughout training; and (2) \textit{Faithful $\to$ Creative}, a reverse schedule that begins with hallucination mitigation and transitions to creative writing. 

Table~\ref{tab:curriculum_learning_strategy} compares the reported curriculum outcomes. Relative to \textit{Random Mixture}, \textit{Creative $\to$ Faithful} lowers the HHEM hallucination rate from 4.09\% to 1.08\% and the MultiHopRAG rate from 1.17\% to 1.02\%, while increasing Creative Writing from 23.65\% to 27.54\%.

The reverse \textit{Faithful $\to$ Creative} schedule obtains the highest Creative Writing score (34.30\%), but has higher hallucination rates on HHEM (9.64\%) and MultiHopRAG (1.72\%) than \textit{Creative $\to$ Faithful} (1.08\% and 1.02\%). Thus, the proposed schedule favors lower HA-GRM-judged hallucination rates, with a lower creative-writing score than the reverse schedule. The preferred schedule depends on the relative priority of these metrics.

\begin{figure}[!ht]
    \centering 
    \includegraphics[width=\linewidth]{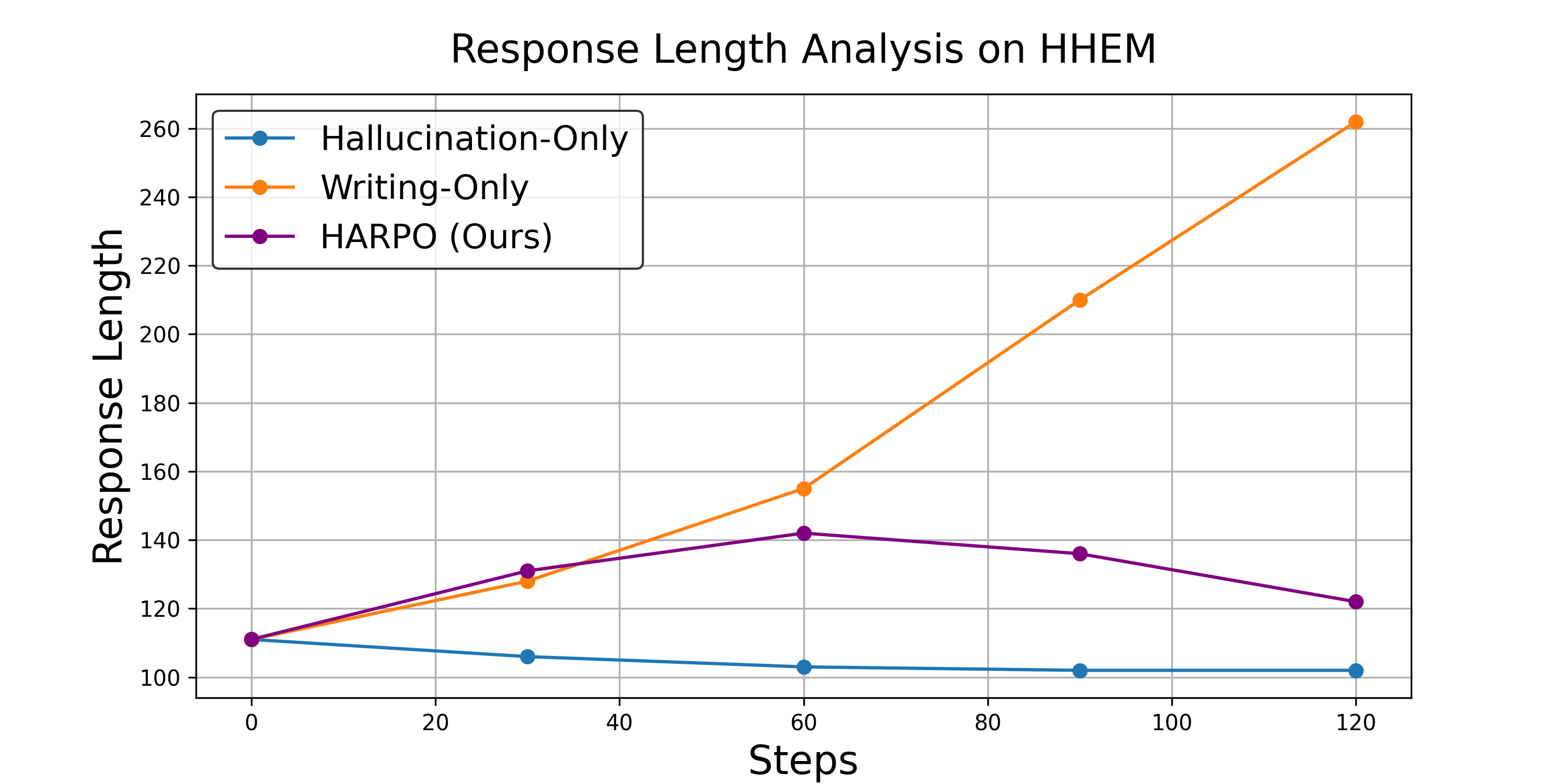} 
    \caption{Response length analysis on HHEM.}
    \label{fig:response_length_analysis_hhem} 
\end{figure}
\paragraph{Analysis of Response Length and Verbosity}
To investigate whether our framework mitigates ``reward hacking'', specifically the tendency of RL-tuned models to inflate scores through verbosity, we track the evolution of response length throughout the training process (Figure~\ref{fig:response_length_analysis_hhem}). The \textit{Writing-Only} baseline exhibits a linear, unbounded increase in token count, confirming that optimizing solely for creativity encourages excessive length. Conversely, the \textit{Hallucination-Only} baseline rapidly converges to shorter, risk-averse responses, potentially sacrificing necessary detail to minimize hallucination rate. HARPO, however, demonstrates a distinct trajectory that balances these extremes. Initially, the response length increases, aligning with the high ratio of creative tasks in the early curriculum. However, as the training progresses beyond step 60, the length peaks and gradually stabilizes. This shift corresponds directly with our curriculum schedule and the imposition of the Selective Activation Mechanism; as the distribution shifts toward faithful tasks and the SAM strictness takes effect, the model is regularized against gratuitous verbosity while retaining sufficient length to remain expressive.

\begin{takeaway}
    {Takeaway (RQ2)}
    {
        HARPO improves the reported hallucination and writing metrics relative to the base models. Compared with Linear Mixture on Qwen3-4B, it has a lower HHEM hallucination rate and higher General score, with a slightly higher MultiHopRAG hallucination rate and lower Creative Writing score.
    }
\end{takeaway}

\section{Related Work}

\paragraph{Closed-Domain Hallucination Detection}
Closed-domain hallucination refers to generating content that is inconsistent with or unsupported by the context or knowledge provided in the input prompt~\cite{Ji2022SurveyOH,Wu2023RAGTruthAH}. Traditional detection methods frame this as a Natural Language Inference (NLI) task~\cite{lattimer2023fast,hhem-2.1-open}, where the source document serves as the premise and the generated text as the hypothesis.
As generative models improve, question answering provides another approach to hallucination detection~\cite{honovich2021q2,cattan2024localizing}.
More recent methods explore intrinsic LLM features, including internal activations and hidden-state dynamics~\cite{ridder2025hallurag,Zhang2025ICRPT}, as well as embedding distances~\cite{ricco2025hallucination}. Complementary resources include HalluLens for benchmarking hallucinations~\cite{bang2025hallulens} and RAGTruth for training and evaluating detectors with response- and span-level annotations~\cite{Wu2023RAGTruthAH}. RL4HS~\citep{su2025learning} trains a reasoning-based hallucination span detector using GRPO with verifiable span-F1 rewards and introduces class-aware policy optimization to address reward imbalance. HARPO combines response- and span-level hallucination evaluation with writing preference evaluation in a generative reward model, then uses its feedback to optimize a generation policy through SAM and a creative-to-faithful data curriculum.

\paragraph{Reinforcement Learning with Verifiable Rewards}
Reinforcement Learning with Verifiable Rewards (RLVR) is a prominent method for enhancing LLM reasoning~\cite{guo2025deepseek}, initially successful in structured domains like mathematics and programming~\cite{wen2025reinforcement} using algorithms such as PPO~\cite{schulman2017proximal} and GRPO~\cite{shao2024deepseekmath}. RL4HS~\citep{su2025learning} extends this approach to hallucination span detection using annotated spans as verifiable supervision. In HARPO, we use verifiable hallucination labels, span annotations, and pairwise preference labels to train HA-GRM; the resulting learned evaluator supplies rewards for downstream policy optimization.

\paragraph{Joint Optimization of Writing Quality and Constraints}
RLMR~\citep{liao2025rlmr} combines subjective writing rewards with objective constraint verification and dynamically adjusts reward weights within sampled groups to penalize constraint violations. HARPO uses predicted context faithfulness to gate the writing reward for each response, together with a curriculum that shifts training from creative to faithful tasks. Our contribution centers on combining hallucination-aware reward modeling with this conditional reward and data schedule to improve the balance between grounded generation and writing quality.

\paragraph{Reducing Hallucination for LLMs}
Research into mitigating LLM hallucinations spans data preparation, training, and inference. RefinedWeb studies large-scale filtering of web pretraining data~\cite{penedo2023refinedweb}, while LIMA studies alignment through supervised fine-tuning on a small curated dataset~\cite{zhou2023lima}. MOSS uses synthetic conversational data for instruction tuning~\cite{sun2023moss}, and Knowledge Consistent Alignment addresses inconsistencies between alignment data and a model's knowledge~\cite{wan2024knowledge}. RLHF aligns models with human preferences~\cite{ouyang2022training}, while process supervision evaluates intermediate reasoning steps~\cite{lightman2023let}. TruthRL examines the trade-off between correct answers, hallucinations, and abstention~\cite{wei2025truthrl}. Factuality Enhanced Language Models use factuality-oriented continued training~\cite{lee2022factuality}. At inference time, layer-contrastive decoding~\cite{chuang2023dola}, context-aware decoding~\cite{shi2024trusting}, and Chain-of-Verification~\cite{dhuliawala2309chain} reduce hallucinations through changes to decoding or response verification.

% !TeX root = ../acl_latex.tex
\section{Conclusion}
% In this work, we presented HARPO, a framework designed to reconcile hallucination mitigation with creative generation. By integrating a Selective Activation Mechanism (SAM) with a dynamic Creative $\to$ Faithful curriculum, HARPO enforces faithfulness as a strict prerequisite for creativity while preventing catastrophic forgetting. Empirical results demonstrate that our approach reduces hallucinations while incentivizing writing quality.
We presented HARPO, a reinforcement learning framework for joint optimization of hallucination mitigation and writing quality. Its Selective Activation Mechanism activates writing rewards only for responses judged hallucination-free by HA-GRM, alongside a curriculum that shifts from creative to faithful tasks. The reported results show lower HA-GRM-judged hallucination rates and higher writing scores relative to the base models. Compared with Linear Mixture on Qwen3-4B, HARPO lowers the summarization hallucination rate and raises the General score, with a slightly higher MultiHopRAG hallucination rate and lower Creative Writing score. The curriculum results likewise show lower hallucination rates than the reverse schedule with a lower creative-writing score. These findings characterize the observed trade-offs across training strategies.

% !TeX root = ../acl_latex.tex
\section*{Limitations}
Despite the promising results of HARPO, several limitations remain. First, training the Hallucination-Aware Generative Reward Model (HA-GRM) relies on high-quality, span-level hallucination annotations, which are significantly more labor-intensive and costly to acquire than standard binary labels. Second, the inference overhead introduced by the generative reward process increases the computational cost of training compared to traditional discriminator-based reward models. Finally, while we demonstrate effectiveness on models up to 8B parameters within general instruction and RAG domains, we have not yet verified the scalability of our framework on significantly larger models (e.g., 70B+) or its transferability to specialized reasoning tasks such as mathematics and code generation, which we leave for future investigation.

SAM depends on HA-GRM's predictions and provides no guarantee of factual correctness. If HA-GRM misses a hallucination, the affected response can still receive a writing reward; if it incorrectly flags a faithful response, that reward is suppressed. These errors can affect policy optimization. Since HA-GRM also judges the reported generation hallucination rates, improvements on that metric should be interpreted with this shared evaluator in mind.

\section*{Ethics Statement}
This study uses existing public datasets for model training and evaluation, including RAGTruth, HaluEval, Arena-Human-Preference-140k, and Arena-Hard-v2.0. The RAGTruth and HaluEval repositories provide MIT licenses.\footnote{\href{https://github.com/ParticleMedia/RAGTruth/blob/main/LICENSE}{RAGTruth license}; \href{https://github.com/RUCAIBox/HaluEval/blob/main/LICENSE}{HaluEval license}.} Arena-Human-Preference-140k licenses user prompts under CC BY 4.0, while model outputs remain subject to the respective providers' terms.\footnote{\href{https://huggingface.co/datasets/lmarena-ai/arena-human-preference-140k}{Arena-Human-Preference-140k dataset card}.} The Arena-Hard-Auto dataset card lists Apache 2.0.\footnote{\href{https://huggingface.co/datasets/lmarena-ai/arena-hard-auto}{Arena-Hard-Auto dataset card}.} Licenses and usage terms for underlying source corpora and pretrained models apply separately.

Public user conversations and preference annotations may contain sensitive information and reflect contributor biases. Reuse and redistribution should avoid disclosing personal information and must retain attribution and applicable usage restrictions.

HA-GRM can miss unsupported claims or incorrectly flag valid content, and these errors can affect both policy optimization and reported hallucination rates. Preference optimization may also favor writing styles represented in the training data. The experiments do not establish safety or reliability for high-stakes applications; deployment requires independent evaluation and appropriate human oversight.

\bibliography{custom}

\appendix

\end{document}